\documentclass{article}

\usepackage[final]{neurips_2022}

\usepackage[utf8]{inputenc} 
\usepackage[T1]{fontenc}    
\usepackage{hyperref}       
\hypersetup{
    colorlinks=true,
    linkcolor=blue,    
    citecolor=blue,     
    urlcolor=blue      
}
\usepackage{url}            
\usepackage{booktabs}       
\usepackage{amsfonts}       
\usepackage{nicefrac}       
\usepackage{microtype}      
\usepackage{xcolor}         
\usepackage{graphicx}
\usepackage{multirow}
\usepackage{pifont}
\usepackage{colortbl}
\usepackage{subcaption}
\usepackage{lipsum} 
\usepackage{natbib}
\setcitestyle{numbers,square}
\bibpunct{[}{]}{,}{n}{,}{,}

\usepackage {ulem}

\title{SAM-Med2D}

\author{
Junlong Cheng\textsuperscript{1,2}\thanks{This work is done when Junlong Cheng is an intern at Shanghai AI Laboratory.}
\qquad Jin Ye\textsuperscript{2}
\qquad Zhongying Deng\textsuperscript{2}
\qquad Jianpin Chen\textsuperscript{2}
\qquad \textbf{Tianbin Li\textsuperscript{2}}
\\
\qquad \textbf{Haoyu Wang\textsuperscript{2}}
\qquad \textbf{Yanzhou Su\textsuperscript{2}}
\qquad \textbf{Ziyan Huang\textsuperscript{2}}
\qquad \textbf{Jilong Chen\textsuperscript{1}}
\qquad \textbf{Lei Jiang\textsuperscript{1}}
\\
\qquad \textbf{Hui Sun\textsuperscript{2}}
\qquad \textbf{Junjun He\textsuperscript{2}}
\qquad \textbf{Shaoting Zhang\textsuperscript{2}}
\qquad \textbf{Min Zhu\textsuperscript{1, $\dagger$}}
\qquad \textbf{Yu Qiao\textsuperscript{2, $\dagger$}}\\
\\
\textsuperscript{1}Sichuan University\qquad \\ \textsuperscript{2}Shanghai AI Laboratory\\ 
{\tt\small {chengjunlong}@scu.stu.edu.cn}\\
{\tt\small \{yejin, hejunjun, litianbin, zhangshaoting, qiaoyu\}@pjlab.org.cn}
}

\begin{document}

\maketitle

\begin{abstract}
The Segment Anything Model (SAM) represents a state-of-the-art research advancement in natural image segmentation, achieving impressive results with input prompts such as points and bounding boxes. 
However, our evaluation and recent research indicate that directly applying the pretrained SAM to medical image segmentation does not yield satisfactory performance. This limitation primarily arises from significant domain gap between natural images and medical images. 
To bridge this gap, we introduce SAM-Med2D, 
the most comprehensive studies on applying SAM to medical 2D images. Its comprehensiveness manifests in three aspects: the comprehensive analysis on collecting the largest medical data, the most comprehensive studies on various fine-tuning options, the most comprehensive evaluation on the performance.
Specifically, we first collect and curate approximately 4.6M images and 19.7M masks from public and private datasets,
constructing a large-scale medical image segmentation dataset encompassing various modalities and objects.
Then, we comprehensively fine-tune SAM on this dataset and turn it into SAM-Med2D. Unlike previous methods that only adopt bounding box or point prompts as interactive segmentation approach, we adapt SAM to medical image segmentation through more comprehensive prompts involving bounding boxes, points, and masks. We additionally fine-tune the encoder and decoder of the original SAM to obtain a well-performed SAM-Med2D, leading to the most comprehensive fine-tuning strategies to date. 
Finally, we conducted a comprehensive evaluation and analysis to investigate the performance of SAM-Med2D in medical image segmentation across various modalities, anatomical structures, and organs. Concurrently, we validated the generalization capability of SAM-Med2D on 9 datasets from MICCAI 2023 challenge. Overall, our approach demonstrated significantly superior performance and generalization capability compared to SAM. Our codes can be found at \url{https://github.com/uni-medical/SAM-Med2D}.
\end{abstract}

\section{Introduction}

Medical image segmentation plays a crucial role in the analysis of medical images by identifying and delineating various tissues, organs, or regions of interest. Accurate segmentation can assist doctors in precisely identifying and locating areas of pathology, enabling more accurate diagnosis and treatment \cite{01}. Furthermore, quantitative and qualitative analysis of medical images provides comprehensive insights into the morphology, structure, and function of different tissues or organs, facilitating disease research and discoveries \cite{02}. However, most existing methods are limited to specific modalities, organs, or lesions due to the characteristics of medical imagery, such as numerous modalities, complex tissue and organ structures, and few annotated data available \cite{03,04,05}. This limitation hinders the generalizability and adaptability of algorithms, making it challenging to apply them across diverse clinical scenarios. 

\begin{figure}[h]
  \centering
  \includegraphics[width=0.95\textwidth]{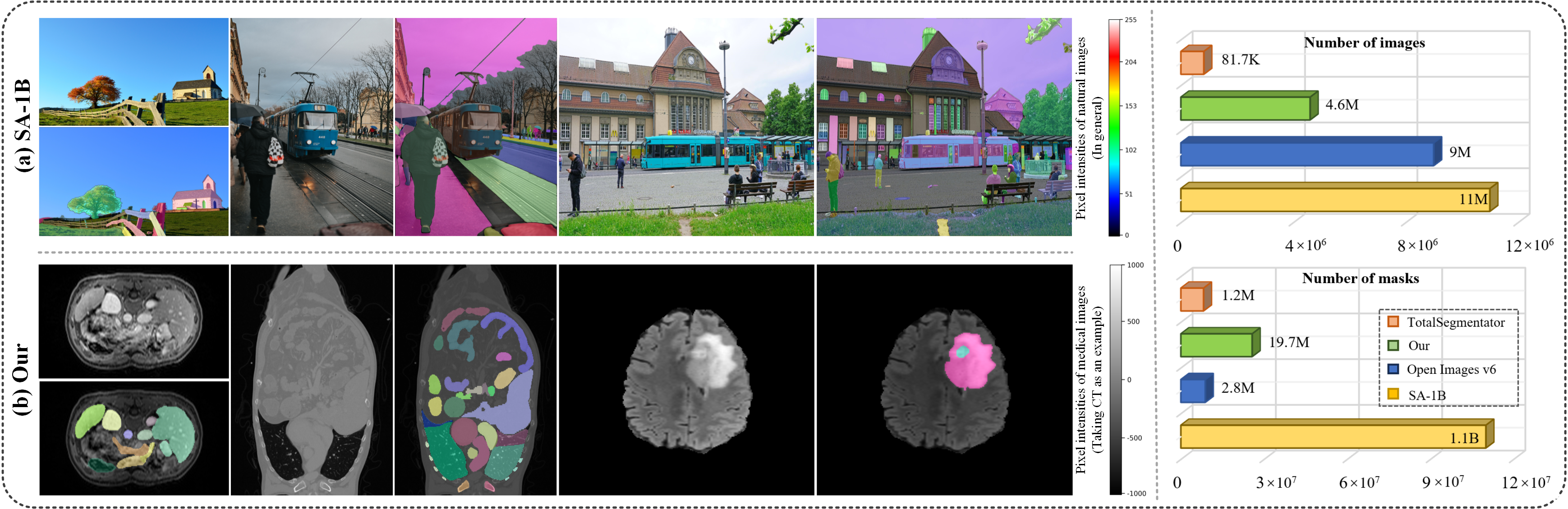}
  \caption{Comparison between examples in SA-1B (a) and in our dataset (b). SA-1B consists of 11M natural images and their corresponding 1129M masks. Our dataset consists of 4.6M medical images and their corresponding 19.7M masks.}\label{fig0}
\end{figure}

Recently, the trend towards large-scale models has created a buzz throughout the AI field. The emergence of general-AI models such as ChatGPT\footnote{https://chat.openai.com}, ERNIE Bot \footnote{https://yiyan.baidu.com/}, DINO \cite{06}, SegGPT \cite{07}, SAM \cite{08}, 
facilitates the use of a single model to address multiple tasks. As the latest large-scale vision model, SAM enables users to generate masks for specific regions of interest through interactive clicking, bounding boxes, or providing natural language prompts. Its zero-shot and few-shot capabilities on natural images \cite{09,10} have garnered significant attention across various domains. 

In the field of medical imaging, some works \cite{11,12,13,14,15} have also focused on the zero-shot capabilities of SAMs. 
However, due to the significant domain gap between natural images and medical images, SAM struggles to generalize to multi-modal and multi-object medical datasets, resulting in unstable segmentation performance across datasets. The reason can be attributed to the data collection methods: medical images are collected from certain protocols and scanners and are presented as different modalities (electrons, lasers, X-rays, ultrasound, nuclear physics, and magnetic resonance) due to their particular clinical purpose. As such, these images are based on a range of physics-based properties and energy sources, which differ greatly from natural images. As shown in Figure~\ref{fig0}, there are significant differences between natural images and medical images in terms of pixel intensity, color, texture, and other distribution characteristics. 
Therefore, the limitation that SAM cannot directly be applied to the medical domain is expected~\cite{11,12,13,14,15}: given that SAM is trained on natural images only, it lacks specific knowledge related to medical imaging.

Equipping SAM with medical knowledge is hard because of the high annotation cost and the diverse annotation quality. Preparing medical data requires domain knowledge and their quality varies significantly across hospitals and clinical studies. These challenges have resulted in a significant disparity between the quantities of medical images and natural images. The bar chart in Figure~\ref{fig0} compares the data volume of publicly available natural image datasets and medical image datasets. For instance, Totalsegmentor, the largest public segmentation dataset in the medical domain, also has a significant gap compared to Open Image v6~\cite{openv6} and SA-1B~\cite{08}.
In this study, our objective is to transfer SAM from natural images to medical images. This will provide benchmark models and evaluation frameworks for researchers in medical image analysis to explore and enhance. To achieve this goal, we proposed SAM-Med2D, the most comprehensive studies on applying SAM to medical 2D images by addressing the following issues:

\begin{figure}[ht]
  \centering
  \includegraphics[width=0.95\textwidth]{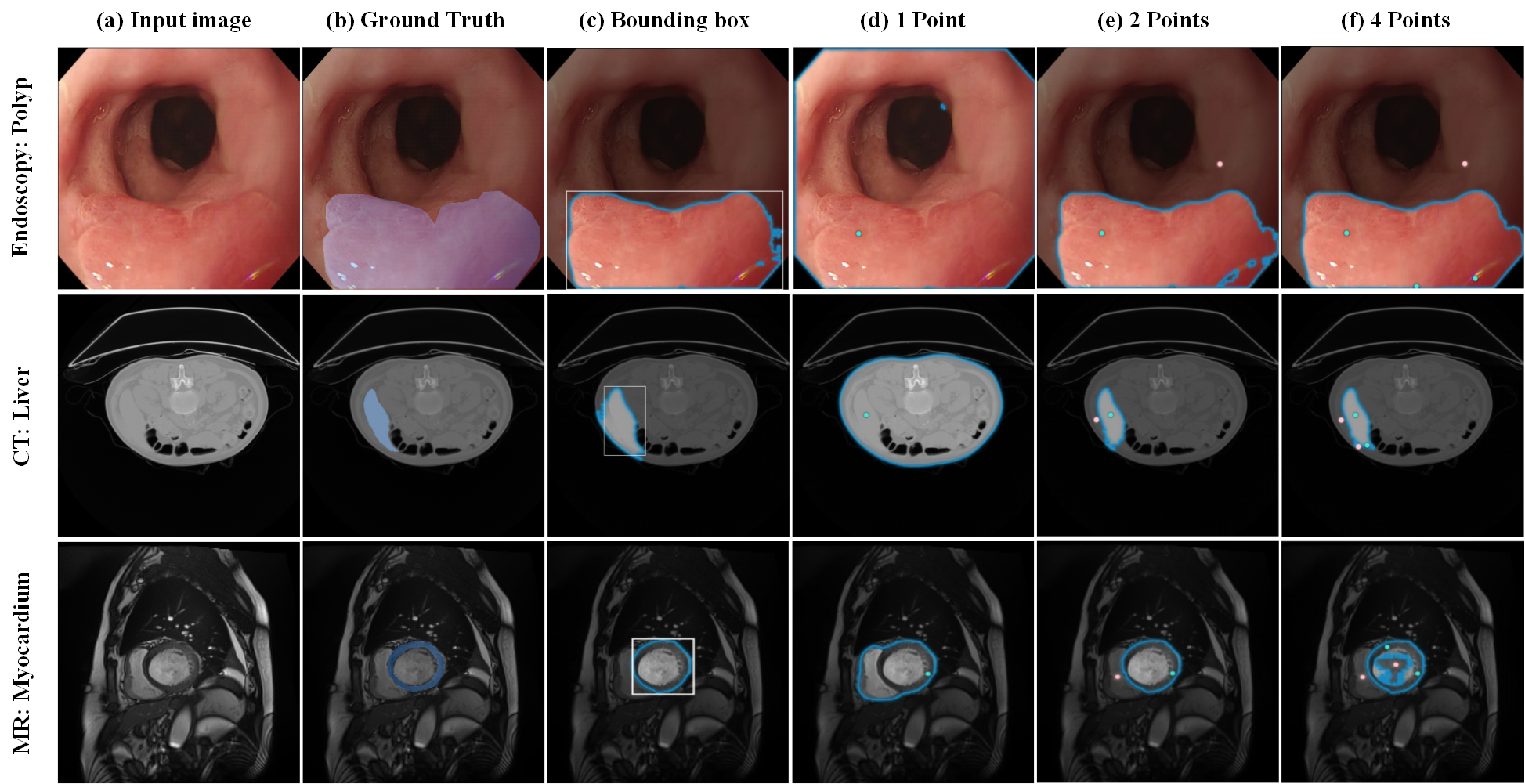}
  \caption{Results of interactive segmentation using SAM in various medical scenarios.}\label{fig1}
\end{figure}

\begin{table}[h]
\renewcommand{\arraystretch}{1.0}
\centering
\caption{Comparison of SAM fine-tuning models. Our SAM-Med2D is a comprehensive fine-tuning method that supports multiple prompts on medical images to generate masks.}\label{tab1}
\begin{tabular}{l|ccccccc}
\toprule
\multirow{2}{*}{\textbf{Model}} & \multirow{2}{*}{\textbf{Dataset (size)}} & \textbf{Encoder}      & \multicolumn{3}{c}{\textbf{Prompts mode}} & \multirow{2}{*}{\textbf{Decoder}} \\ \cline{4-6}
                       &                               & (or Adapter) & \textit{Point}      & \textit{Bbox}     & \textit{Mask}     &                          \\ \hline
SAM-U \cite{16}                   & 6000 masks                     & \textcolor{red}{\ding{56}}    & \textcolor{red}{\ding{56}}  & \textcolor{green}{\ding{52}} & \textcolor{red}{\ding{56}} & \textcolor{red}{\ding{56}}                         \\ 
SAMed \cite{17}                  & 3779 masks                      & \textcolor{green}{\ding{52}}  & \textcolor{red}{\ding{56}}  & \textcolor{red}{\ding{56}}   & \textcolor{red}{\ding{56}} & \textcolor{green}{\ding{52}}                         \\
AutoSAM \cite{18}                & ACDC \cite{19}                  & \textcolor{red}{\ding{56}}    & \textcolor{red}{\ding{56}}  & \textcolor{red}{\ding{56}}   & \textcolor{red}{\ding{56}} & \textcolor{green}{\ding{52}}                         \\
MedSAM \cite{20}                  & $\sim$1.1M masks                & \textcolor{red}{\ding{56}}    & \textcolor{red}{\ding{56}}  & \textcolor{green}{\ding{52}} & \textcolor{red}{\ding{56}} & \textcolor{green}{\ding{52}}                         \\
MSA \cite{21}                    & 5 datasets                   & \textcolor{green}{\ding{52}}    & \textcolor{green}{\ding{52}}& \textcolor{red}{\ding{56}}   & \textcolor{red}{\ding{56}} & \textcolor{green}{\ding{52}}                          \\ \hline
SAM-Med2D (Ours)          & $\sim$19.7M masks                       & \textcolor{green}{\ding{52}}    & \textcolor{green}{\ding{52}}& \textcolor{green}{\ding{52}} & \textcolor{green}{\ding{52}}& \textcolor{green}{\ding{52}}                         \\ 
\bottomrule

\end{tabular}
\end{table}

\textbf{\textbullet\quad How to fine-tune SAM for the medical imaging domain?}

1) We needed to incorporate knowledge about medical images into SAM, so we collected and curated a medical image segmentation dataset comprising over 4.6M images and 19.7M masks. To the best of our knowledge, this dataset represents \textit{the largest medical image segmentation dataset}, encompassing multiple modalities and covering comprehensive objects. Table~\ref{tab1} illustrates the methods for fine-tuning SAM on specific limited-scale medical datasets. While these methods have proven effectiveness, they only enhance SAM's segmentation capabilities within specific scenarios resembling the training dataset. As a result, their applicability is limited for more diverse medical image segmentation tasks.

2) Different prompt modes play important roles in different segmentation tasks when migrating SAM to the medical field. As shown in Figure ~\ref{fig1} (c), a relatively accurate polyp mask can be obtained by using the bounding box prompt. In contrast, the mask quality is poor when clicking on a foreground point (column d). With an increasing number of clicks, the segmentation result gradually improves and even surpasses the performance of the bounding box prompt (e.g., liver segmentation). When segmenting myocardium, using the bounding box prompt may include uninterested regions in the result, while the point prompt allows us to gradually acquire the desired mask. Therefore, this paper aims to fine-tune three prompt modes (point, bounding box, and mask) to meet the requirements of different scenarios in medical image segmentation tasks. According to Table ~\ref{tab1}, our approach involves a more \textit{comprehensive fine-tuning} compared to other methods, which means our method possesses a broader range of prompt segmentation capabilities in the medical image domain.

3) Adapters have been proven to be an effective strategy for fine-tuning large-scale models \cite{21,chen2022vitadapter}. They do not require retraining the entire model, ensuring that the original knowledge is not forgotten. 
This enables an existing model to excel in both new and the original tasks. Due to the parameter-sharing property of adapters, they facilitate transfer learning between different tasks. Therefore, this paper applies Adapter to the encoder of SAM to learn domain-specific information in the field of medical imaging. Additionally, the adapter layer is a plug-and-play component that allows us to choose whether to keep or remove it during the testing process, according to our specific needs.

\textbf{\textbullet\quad How to evaluate the performance of SAM-Med2D?}

To comprehensively evaluate the performance of SAM-Med2D in the medical imaging domain, we assessed its capabilities from multiple perspectives. We focused on the following aspects: 
\textbf{1) Data diversity:} We evaluated SAM-Med2D on various imaging modalities such as CT, MR, X-ray, different organs, and multiple pathological conditions like tumors, inflammations, etc., to ensure its ability to segment different types of medical images.
\textbf{2) Fine-tuning strategy:} We compare the default fine-tuning strategy of SAM-Med2D with other alternative strategies, e.g., different model configurations, to demonstrate SAM-Med2D's success in the medical imaging domain.
\textbf{3) Generalization ability:} We evaluated the robustness of SAM-Med2D using 9 MICCAI2023 datasets to ensure its accurate segmentation in unseen medical image environments.

Through \textit{comprehensive evaluation}, we found that SAM-Med2D possesses the following capabilities: 
\textbf{1)  Medical expertise:}  Compared to SAM, SAM-Med2D demonstrates superior performance in handling complex organ structures, lesions, and cases with unclear boundaries. This means that SAM-Med2D can accurately identify and segment challenging areas in medical images, thus providing more precise diagnostic and treatment support.
\textbf{2) Broad segmentation capabilities:} SAM-Med2D demonstrates a broad segmentation capability across various prompt modes, enabling it to fulfill segmentation tasks in different scenarios. This implies that doctors and medical imaging professionals can utilize SAM-Med2D to perform more precise and accurate segmentation operations, thereby enhancing the efficiency and reliability of medical image analysis results.
\textbf{3) Generalization ability:} SAM-Med2D demonstrates strong generalization ability, allowing direct application to unseen medical image data and producing good segmentation results.

\section{Related Work}
\textbf{Large-scale Vision Models (LVM).} 
Inspired by large language models such as ChatGPT and GPT-4\footnote{https://openai.com/research/gpt-4}, researchers have developed similar LVM including \cite{07,08,22,23,24,25,26}. These models exhibit outstanding zero-shot and few-shot generalization capabilities, enabling rapid adaptation and extension to target tasks or domains through pre-training and fine-tuning paradigms. Among them, CLIP \cite{23} provides a unified vision and language model that can be utilized for various tasks, including classification, detection, and visual question answering. Through extensive pre-training on text and image pairs, this model has achieved impressive results in multiple benchmark tests. DALL· E \cite{25} is a variant of the Large-Scale Transformer model, GPT-3, trained to generate images based on textual descriptions. Recently, SAM \cite{08} pre-trained on 1B masks has become a general-purpose LVM for image segmentation. It demonstrates powerful zero-shot capabilities, allowing interactive or automatic segmentation of any object. In contrast, SegGPT \cite{07} unifies different segmentation tasks into a single contextual learning framework by transforming diverse segmentation data into images of a standardized format. Furthermore, SEEM \cite{26} proposes a universal interface that employs multi-modal prompts to simultaneously segment all contents in images or videos and identify object categories. However, these LVMs have not been explicitly optimized for the field of Medical Image Analysis (MIA).

\textbf{Fine-tuned SAM in Medical Image Analysis.} 
SAM provides an excellent framework for interactive segmentation, making it a benchmark model for prompt-based medical image segmentation. However, due to significant domain differences between natural images and medical images, SAM's performance significantly declines when applied to medical images \cite{16,17,18,20,21}. Current research primarily focuses on fine-tuning SAM for specific medical segmentation datasets. Deng et al. \cite{16} proposed a multi-box prompt triggering uncertainty estimation for SAM, achieving significant improvements in retinal image segmentation. Zhang et al. \cite{17} applied a low-rank-based fine-tuning strategy to the SAM encoder while simultaneously fine-tuning the decoder for abdominal segmentation tasks. Hu et al. \cite{18} abandoned the original prompt encoder of SAM and constructed different types of prediction heads for fine-tuning to accomplish automated cardiac segmentation. Ma et al. \cite{20} collected 11 different modalities of medical image data and fine-tuned the mask decoder of SAM on over 1M masks while preserving the original box prompt. In contrast to the aforementioned studies, Wu et al. \cite{21} proposed the MSA, which integrates medical-specific domain knowledge into SAM using a simple adapter technique and validated it on 19 medical image segmentation tasks. These studies demonstrate that fine-tuning or adapter techniques can improve SAM's performance in medical image segmentation. 
Unlike the aforementioned methods that only provide bounding box or point prompt, we adapt SAM to medical image segmentation by using more comprehensive prompts (i.e., bounding box, points, and mask). In addition, we comprehensively analyze the performance and challenges of our method in medical image segmentation by considering different imaging modalities, anatomical structures, and organs in medical images.

\textbf{Zero-shot evaluation of SAM in medical imaging.} 
Recent studies have reported the zero-shot performance of SAM in medical image segmentation. Deng et al. \cite{27} investigated SAM's segmentation capabilities on tumor and tissue tasks under different prompt conditions, and the experimental results show that SAM performs better only on segmenting large connected objects. Hu et al. \cite{28} evaluated the impact of point prompts on SAM's performance in multi-phase liver tumor segmentation in CT volumes. The results demonstrated that SAM's performance improved as the number of point prompts increased. Zhou et al. \cite{12} tested SAM's performance in colonoscopy polyp segmentation without prompts, and the results indicated significant room for improvement when applying SAM to polyp segmentation tasks. Cheng et al. \cite{29} extensively evaluated the following models on 12 open-source medical image datasets: promptless models, models with 1, 3, and 10-point prompts, and models with box prompts with 5 different jitter levels. The performance of SAM was generally lower compared to state-of-the-art results. Similarly, Huang et al. \cite{15} evaluated the zero-shot performance of SAM on 52 public datasets using three different prompts, and the evaluation results consistently showed unsatisfactory performance of SAM across various medical image segmentation tasks. 
We believe that analyzing SAM's performance on large-scale medical image datasets is crucial. This can help the community gain a better understanding of the factors influencing the model's ability to perceive medical objects. These factors can contribute to the improved development of general medical segmentation methods. Therefore, this study provides a comprehensive evaluation of SAM-Med2D from multiple perspectives.

\begin{figure}[ht]
  \centering
  \includegraphics[width=1.0\textwidth]{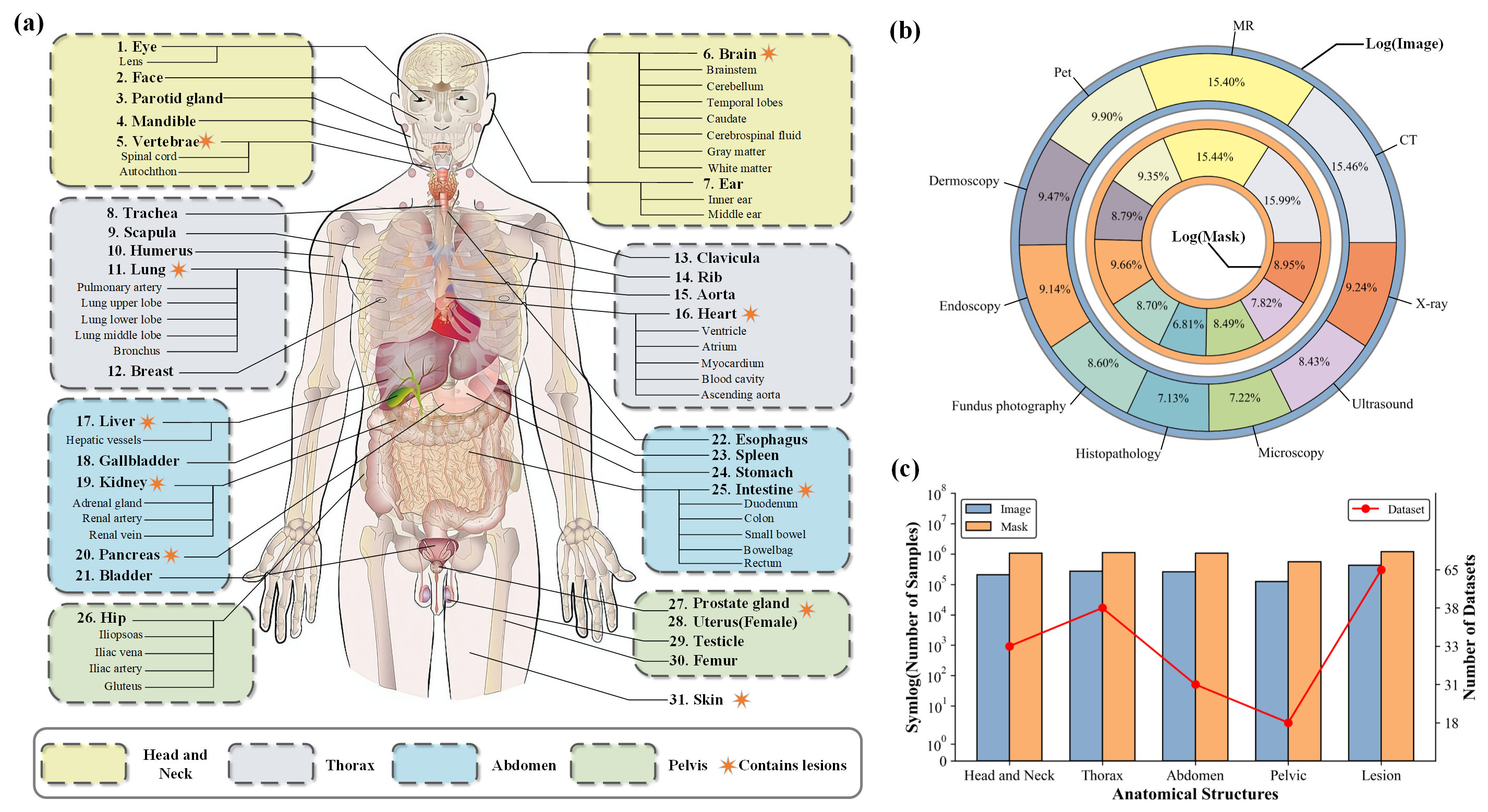}
  \caption{Overview of the dataset used in this study. (a) A total of 31 major organs, along with their corresponding anatomical structures, with an asterisk (*) denoting the presence of lesion labels within the dataset. (b) The distribution of modalities along with their corresponding proportions in the dataset are presented (scaled logarithmically). (c) The number of images and masks categorized by anatomical structure, along with the total count encompassing the dataset.}\label{fig2}
\end{figure}

\section{Methods}

\subsection{Incorporation of Medical Knowledge into SAM}

Recent research has reaffirmed the pivotal role of training data volume in the learning capacity of large models \cite{07,08,23}. By learning from larger-scale data, models can acquire richer domain-specific knowledge and adapt better to various application scenarios. Though trained on over 1B masks, SAM achieves suboptimal performance in the realm of medical image analysis due to the significant domain gap between natural images and medical data. 
To address this gap, we have collected and curated the largest medical image segmentation dataset to date. This dataset is composed of numerous public and private datasets, ensuring comprehensive coverage and diversity. Figure~\ref{fig2} (b) illustrates the dataset's 10 different imaging modalities and their corresponding data proportions. To enhance visual presentation, we have used logarithmic scaling to visualize the differences in quantity. Based on anatomical structures and the presence of lesions, we categorized the dataset into head and neck, thorax, abdomen, pelvic, and lesions (Figure~\ref{fig2} (c)). Additionally, we curated and consolidated 31 main organs from the 271 labels in these datasets, as depicted in Figure~\ref{fig2} (a). This covers almost all object types in the currently available public datasets, addressing the deficiency of SAM in medical domain knowledge.

For the effective application of SAM to medical image segmentation, we preprocessed the dataset from multiple angles. Firstly, for 3D datasets, we normalized the intensity values of each volume to the range [0, 255] and extracted all slice images and their corresponding masks along the x, y, and z axes. In the extraction process, slice images with the shortest edge less than half the length of the longest edge were discarded to prevent target areas from becoming extremely blurry when adjusting for images with large aspect ratios. For 2D datasets, we only checked whether pixel values were within the range [0, 255], and all processed images were saved in PNG format to maintain data loading consistency. Secondly, when a mask contained multiple classes, we generated multiple masks, each containing only one class (similar to sam1B \cite{08}). We also split masks with multiple connected components (for example, left and right lungs) into multiple masks with single connected components. If multiple organs were present and only contained one connected component, we retained the mask to increase data diversity. Lastly, we excluded masks where the target area constituted less than 0.153\% ($\frac{100}{256 \times 256}$) of the total image, meaning that when an image is resized to 256×256, its target area must exceed 100 pixels.

Following these procedures, we obtained approximately 4.6M images and 19.7M masks. We randomly split 80\% data for training and 20\% for testing based on image indices. The resulting training set included roughly 3.67M images and 15.8M masks, while the test set contained 0.92M images and 3.9M masks. We also introduced 9 MICCAI2023 datasets (comprising about 0.52M images and 1.31M masks) that served solely to validate the model's generalization ability. We believe that with a more comprehensive and diverse set of training data, SAM will better adapt to the complexities and nuances of the medical imaging field, providing more accurate and reliable support for applications in healthcare. This will also usher in new opportunities and challenges for research and development in the field of medical image segmentation.

\begin{figure}[ht]
  \centering
  \includegraphics[width=0.95\textwidth]{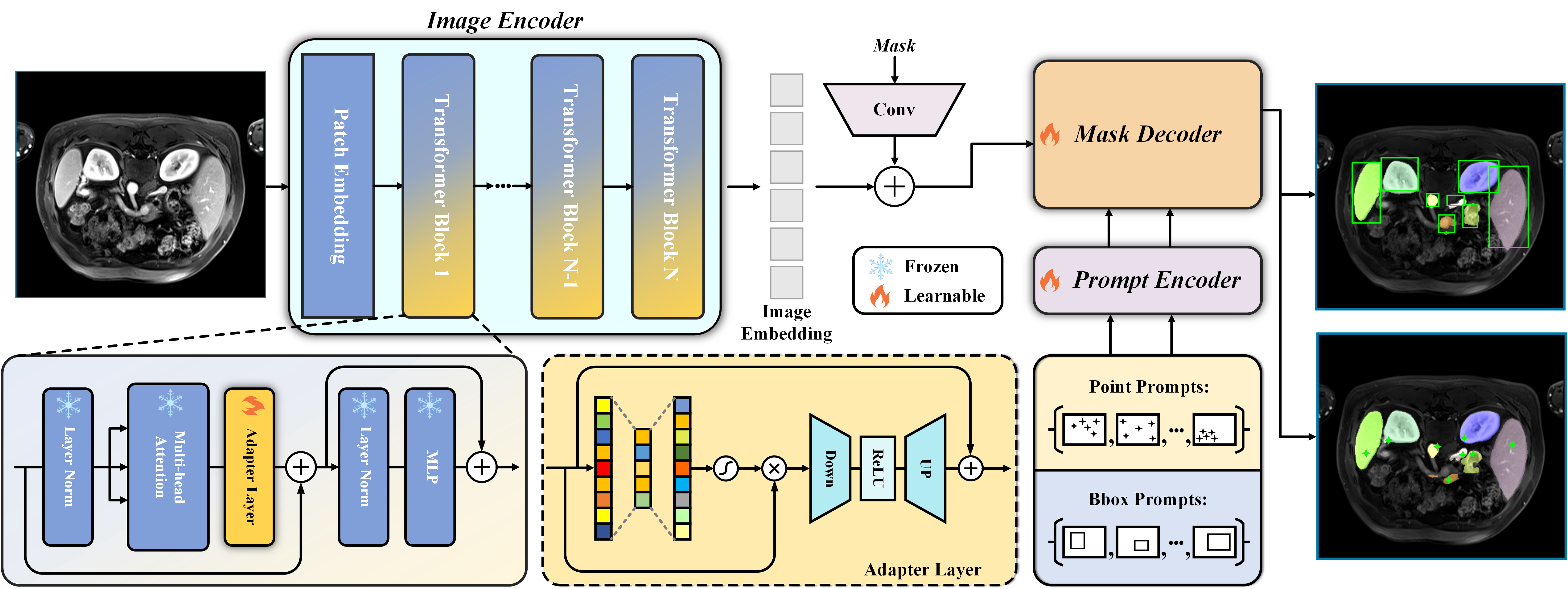}
  \caption{The pipeline of SAM-Med2D. We freeze the image encoder and incorporate learnable adapter layers in each Transformer block to acquire domain-specific knowledge in the medical field. We fine-tune the prompt encoder using point, Bbox, and mask information, while updating the parameters of the mask decoder through interactive training.}\label{fig6}
\end{figure}

\subsection{Transition from SAM to SAM-Med2D}

Before introducing SAM-Med2D, let us briefly review the SAM architecture. SAM consists of three main components: a large-scale image encoder, a prompt encoder, and a lightweight mask decoder. This framework allows for different masks to be generated for the same image based on different prompts. The image encoder utilizes a pre-trained Visual Transformer (ViT) \cite{22} to process high-resolution inputs and outputs feature maps at a 1/16 scale of the original image. The prompt encoder includes sparse prompts and dense prompts, mapping points, bounding boxes, or text to 256-dimensional vectors and applying convolutional down-sampling to dense prompts with GELU activation function at each level. The mask decoder receives embedded information from both encoders, and updates the image embeddings and prompt embeddings through a cross-attention mechanism. In this work, we fine-tune SAM to create SAM-Med2D, which effectively extends the framework to the domain of medical images. We will now discuss each component of SAM-Med2D and the fine-tuning strategies in detail.

\textbf{A. Adapting Image Encoder}

As the most parameter-heavy part of SAM, globally updating the image encoder during fine-tuning incurs significant computational costs. To incorporate medical domain knowledge into the image encoder at a lower cost, we introduce Adapter technology. Specifically, we freeze all parameters of the original image encoder during fine-tuning and deploy an adapter for each Transformer block, as shown in Figure~\ref{fig6}. We adapt the image encoder along both channel and spatial dimensions. For the channel dimension, we first compress the resolution of the input feature map to C×1×1 using global average pooling. Then, we employ a linear layer to compress the channel embeddings and another linear layer to restore them, with a compression ratio of 0.25. Finally, we obtain the weights of the channel dimension through a sigmoid function and multiply them with the input feature map as the input for the next level. For the spatial dimension, we downsample the spatial resolution of the feature map by a factor of two using a convolutional layer and restore the spatial resolution using a transposed convolution, maintaining the same number of channels as the input. A skip connection is added after each adapter layer.

\textbf{B. Prompt Encoder and Mask Decoder}

The prompt encoder in SAM supports four types of prompts: point, bounding box, mask, and text prompts. Due to the lack of large-scale pre-trained models for medical image-text alignment, the use of text prompts is limited. Therefore, we only consider the remaining three prompt modes for fine-tuning. Compared to previous methods that fine-tune a single prompt only \cite{20,21}, we retain the full functionality of prompts and enhance their applicability in the medical imaging domain. Specifically, SAM-Med2D utilizes both sparse prompts (point and bounding box) and dense prompts (mask) simultaneously. For sparse prompts, each point is represented as a vector embedding of its positional encoding along with the sum of two learned embeddings indicating its foreground or background position. Each bounding box uses the positional encoding of its top-left and bottom-right corners, along with learned embeddings representing "top-left" and "bottom-right" as vector embeddings. For dense prompts, we use the low-resolution feature map generated after the first iteration of the model as the mask prompt, applying two convolutional embeddings that reduce the scale of the input mask by a factor of 4 with output channels of 1/4 and 1/16 of the original input. Finally, a 1×1 convolution is used to map the channel dimension to 256.

We did not make any changes to the mask decoder structure and kept updating its parameters during training. To make the model ambiguity-aware, each prompt predicts multiple masks simultaneously (defaulting to three). During backpropagation, we choose the predicted mask with the highest Intersection over Union (IoU) score relative to the ground truth to calculate the loss, propagating gradients accordingly. We map the low-resolution feature map generated from the previous iteration to the range of [0, 1] as the dense prompt for the current iteration. During the actual training process, we observed that even with only sparse prompts, the model can quickly converge, which diminishes the impact of dense prompts. Therefore, we adopt the training strategy of SAM, where in the last iteration and one random intermediate iteration, we only provide dense prompts to encourage the model to benefit from the provided masks.

\textbf{C. Fine-tuning Strategy}

Similar to SAM and other interactive segmentation methods \cite{30}, we train SAM-Med2D by simulating interactive segmentation. For each batch of data, we train the model for 9 iterations. In the first iteration, we randomly select a foreground point or bounding box as the sparse prompt with equal probabilities. The foreground point is sampled from the ground truth, and the bounding box is the maximum enclosing rectangle of the ground truth, with each coordinate offset by a maximum of five pixels. It is worth noting that except for the first iteration, where the parameters of the Adapter layer, prompt encoder, and mask decoder are updated simultaneously, subsequent iterations only update the parameters of the mask decoder. Starting from the second iteration, we randomly select 1, 3, 5, or 9 points from the error region between the previous mask predictions and the ground truth as subsequent sparse prompts, allowing users to perform single-point or multi-point interactive segmentation. SAM-Med2D aims to improve the segmentation of medical images by adapting the SAM framework specifically for the medical imaging domain. It incorporates adapter technology, extends the functionality of prompts, and uses a fine-tuning strategy based on simulated interactive segmentation.

\subsection{Evaluate SAM-Med2D}
Comprehensive performance evaluation is crucial for the research community to gain a deeper understanding of the factors influencing the algorithm's ability to perceive medical objects, thereby improving methods and enhancing their effectiveness in practical applications. However, previous evaluations \cite{11,12,13,15} have been limited by the scarcity of data and the lack of a benchmark for general medical image segmentation methods, resulting in evaluations confined to small-scale datasets and limited categories, failing to fully reveal the strengths and applicability of algorithms. To address this research gap, we will conduct a comprehensive and multidimensional evaluation of SAM-Med2D, providing a benchmark for interactive segmentation methods for future researchers.

In terms of models, we take SAM as the baseline model and select two intuitive prompt modes, Bbox and Points, from its interaction methods to evaluate SAM-Med2D. This choice was made because Bbox and points are commonly used interaction modes and can serve as simple and efficient annotation methods. By evaluating the performance of SAM-Med2D under these two interaction modes, we can delve into their advantages and limitations in medical image segmentation tasks. The Bbox interaction mode guides the algorithm to perform segmentation by bounding the target with a box, which is intuitive and easy to understand. It provides rough positional and shape information of the target, thereby guiding the algorithm to perform more accurate segmentation. However, in some cases, accurately enclosing the target with a Bbox may be challenging, especially for complex-shaped targets, targets with blurred edges, or overlapping objects. This may result in inaccuracies in segmentation results and instances of missed segmentation. On the other hand, the points interaction mode guides the algorithm to perform segmentation by marking key points or regions of the target. This mode can provide more precise segmentation guidance, particularly for targets with complex shapes or local details. These evaluations will help us gain a deeper understanding of the strengths and limitations of different interaction modes in medical image segmentation, thereby inspiring researchers to design and develop more flexible and efficient interaction patterns.

In terms of data, we will evaluate the performance of SAM-Med2D on 10 different modalities of medical images, including MRI, CT, ultrasound, and more. This comprehensive evaluation will allow us to understand the strengths and challenges of SAM-Med2D under specific modalities and reveal its potential application capabilities in multimodal images. Different medical image modalities have distinct characteristics and sources of noise, so evaluating SAM-Med2D's performance on these modalities will provide a more comprehensive understanding of its applicability and robustness. Furthermore, considering that different anatomical structures and organs have unique morphologies, features, and variation patterns, we conducted evaluations of SAM-Med2D on four anatomical structures and 31 major organs. Such evaluations help us gain in-depth insights into the performance differences of SAM-Med2D in different scenarios and enable targeted improvements to address challenges specific to certain structures and organs. Lastly, we place great emphasis on the generalization ability of SAM-Med2D and therefore tested it on 9 MICCAI 2023 datasets. These datasets represent medical images collected from different sources, institutions, or devices, providing diversity. By evaluating SAM-Med2D on these datasets, we can verify its ability to generalize to new data. This is crucial in validating the applicability of our method to a wide range of clinical scenarios and data sources. 

Through the above comprehensive evaluation, we will be able to provide in-depth insights into the performance and applicability of SAM-Med2D, serving as a valuable reference and benchmark for future researchers and developers. These evaluation results will have a positive impact on the development of medical image segmentation, promoting the design and application of more accurate and efficient methods.

\section{Experiments and Evaluation}

\subsection{Implementation Details}

Our method is implemented in PyTorch and trained on 8 NVIDIA Tesla A100 GPUs, each with 80GB memory. Considering the memory constraints, we only fine-tune the base model of SAM (SAM-B) in this work. We use the Adam optimizer with an initial learning rate of 1e-4 and train for a total of 12 epochs, with the learning rate divided by 2 at the 7th and 10th epochs. During training, all images are resized to a resolution of 256x256. Our resizing strategy involves padding the edges with zeros for images with both width and height smaller than 256, while using bilinear interpolation to resize images in other cases. For each image, we randomly select 5 corresponding masks. If there are fewer masks available, we randomly duplicate samples. To make full use of GPU memory, each GPU processes 50 images along with their corresponding 250 masks. The loss function supervising the mask predictions is a linear combination of focal loss \cite{31} and dice loss \cite{32}, with a ratio of 20:1. Additionally, the mean squared error loss between the intersection over union (IoU) prediction and the predicted mask with ground truth mask is used for training. We use the Dice score to evaluate the segmentation results.

\begin{table}[h]
\renewcommand{\arraystretch}{1.1}
\centering
\caption{Quantitative comparison of different methods on the test set.}\label{tab2}
\begin{tabular}{l|lccccc}
\toprule
\multirow{2}{*}{\textbf{Model}} & \multirow{2}{*}{\textbf{Resolution}} & \multicolumn{4}{c}{\textbf{Prompt mode (\%)}} &\multirow{2}{*}{\textbf{FPS}}\\ \cline{3-6} 
                       &                       & \textit{Bbox}    & \textit{1 pt}    & \textit{3 pts}   & \textit{5 pts}  \\ \hline
SAM \cite{08}      & $256\times256$                 & 61.63   & 18.94   & 28.28   & 37.47   & 51        \\
SAM \cite{08}      & $1024\times1024$                  & 74.49   & 36.88   & 42.00   & 47.57  & 8       \\
FT-SAM                 & $256\times256$                & 73.56   & 60.11   & 70.95   & 75.51   & 51       \\
SAM-Med2D                & $256\times256$                  & 79.30   & 70.01   & 76.35   & 78.68  & 35     \\ 
\bottomrule
\end{tabular}
\end{table}

\subsection{Quantitative Evaluation}

\textbf{A. Overall Performance.} Table~\ref{tab2} presents the overall performance results of SAM, FT-SAM (only fine-tune the mask decoder), and our SAM-Med2D on test set. We found that in the bounding box prompt (Bbox prompt) mode, FT-SAM achieved an 11.93\% improvement in Dice score over SAM, while our SAM-Med2D achieved a more significant performance boost, with  Dice scores of 79.30\% (i.e., 17.67\% boost). This indicates that fine-tuning on a large-scale dataset can lead to better transferability in the target domain. We also simulated interactive segmentation in the point prompt mode. In this mode, we randomly sampled a point from the foreground as the first prompt point, and subsequent prompt points were randomly selected within the error region between the segmentation result and the ground truth. Additionally, the low-resolution mask generated in the previous iteration was used as the mask prompt, along with the previous prompt points, as input to the model. Experimental results showed that SAM performed poorly when using a single point prompt, even with a resolution of 1024×1024, exhibiting a 23.23\% lower Dice score compared to FT-SAM. As the number of prompt points increased, the performance of different models significantly improved, with the fine-tuned approach even surpassing the Bbox prompt mode. This demonstrates the feasibility and effectiveness of using point-based interactive segmentation in medical images. Furthermore, the overall segmentation performance of SAM at a resolution of 1024×1024 was inferior to that of the fine-tuned approach. This indicates that the fine-tuned model learned specific knowledge of the medical domain, and fine-tuning with low cost is an effective and feasible method to reduce domain differences.

\begin{figure}
  \centering
  \includegraphics[width=1.0\textwidth]{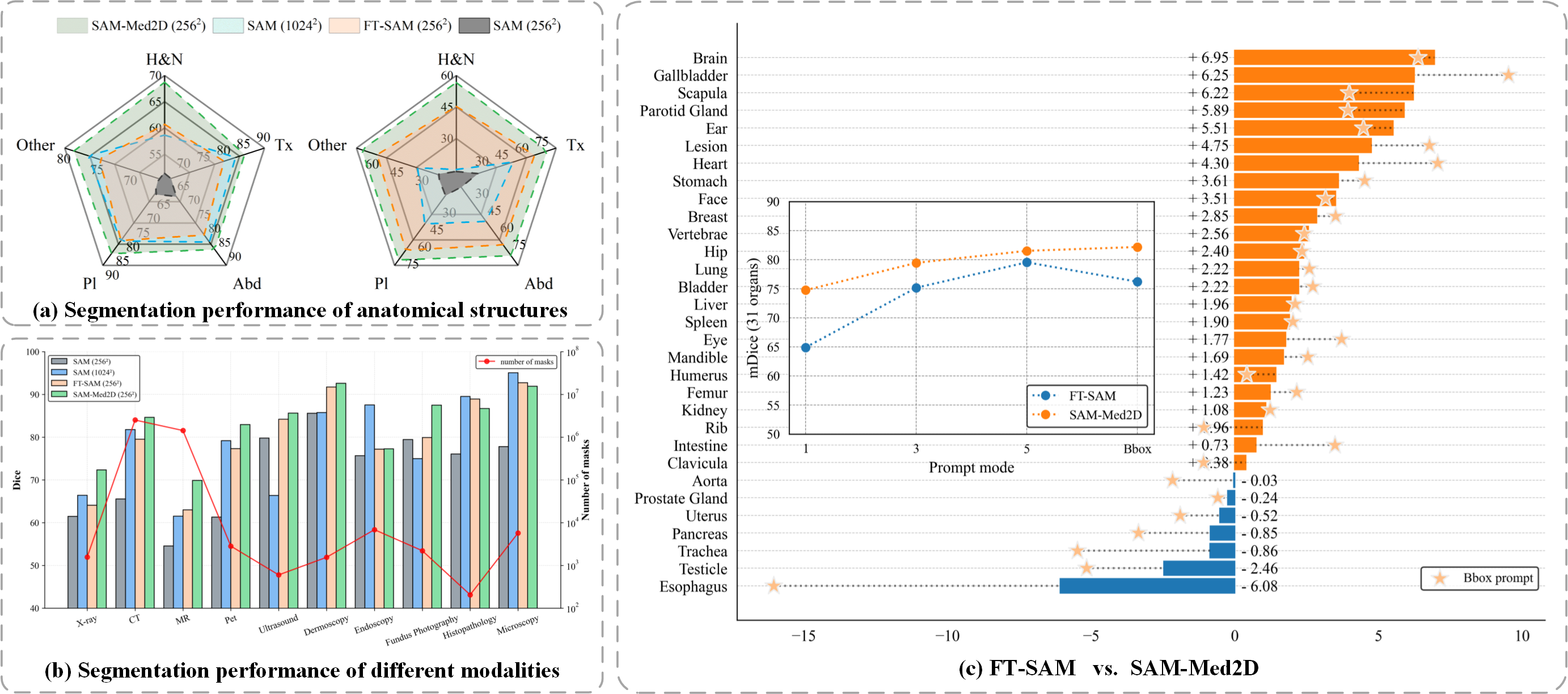}
  \caption{(a) Comparison from the perspective of anatomical structures. (b) Comparison from the perspective of different Modalities. (c) Comparison of segmentation performance between FT-SAM and our SAM-Med2D across 31 organs.}\label{fig3}
\end{figure}

\textbf{B. Performance Evaluation on Anatomical Structures.} As shown in Figure~\ref{fig3} (a), we evaluated the segmentation performance of different models and resolutions on the head and neck (H\&N), thorax (Tx), abdomen (Abd), pelvis (Pl), and other regions. The "other regions" includes lesions and cases outside the aforementioned four anatomical structures. Our main focus was on the segmentation performance of the models using Bbox prompt and 1 point prompt (1 pt prompt). 

We observed that when using the Bbox prompt, SAM (at 1024 ×1024 resolution) outperformed FT-SAM in the Tx, Abd, and other regions. However, it showed poorer performance in the H\&N region. This can be attributed to the relatively smaller size of lesions or organs in the H\&N area, as well as the presence of less clear boundaries, making it challenging for the model to adapt to this type of segmentation task without fine-tuning. Our SAM-Med2D demonstrated advantages in Dice scores across all anatomical structures compared to other methods. Due to the limited information provided by a 1 pt prompt, there were performance differences across different categories. Interesting, we found that the fine-tuned SAM significantly outperformed the original SAM. This is because the fine-tuning approach learned the positional relationships of points within the target regions from a large-scale medical image dataset, enabling more accurate decision-making.

Based on the aforementioned results, we draw the conclusion that SAM-Med2D exhibits excellent performance in the segmentation tasks across different anatomical structures, yielding satisfactory results in terms of the Dice metric in the pelvic and thoracic areas. However, it is worth noting that the performance of the head and neck region appears to be relatively subpar across different models and resolutions, suggesting the need for additional improvement measures.

\begin{table}[ht]
\centering
\caption{Segmentation performance in point prompt mode. The left values represent Dice scores of different models under 1 pt prompt. The numbers in parentheses indicate the Dice score increment after 5 pts prompt, with red indicating improvement and green indicating decline.}\label{tab3}
\small
\begin{tabular}{l|cccc}
\toprule
\multirow{2}{*}{\textbf{Modal}} & \textbf{SAM} \cite{08}        & \textbf{SAM} \cite{08}        & \textbf{FT-SAM}         & \textbf{SAM-Med2D}        \\
                       & (\textit{$256\times256$})         & (\textit{$1024\times1024$})        & (\textit{$256\times256$})         & (\textit{$256\times256$})         \\ \hline
CT                     & 20.87($\Delta \textcolor{red}{18.62}$) & 48.36($\Delta \textcolor{red}{12.74}$) & 67.91($\Delta \textcolor{red}{13.81}$) & 77.34($\Delta \textcolor{red}{6.75}$)  \\
MR                     & 15.25($\Delta \textcolor{red}{18.41}$) & 16.45($\Delta \textcolor{red}{7.07}$)  & 46.36($\Delta \textcolor{red}{18.17}$) & 57.16($\Delta \textcolor{red}{12.02}$) \\
PET                    & 15.12($\Delta \textcolor{red}{25.09}$) & 34.52($\Delta \textcolor{red}{8.93}$)  & 59.58($\Delta \textcolor{red}{11.58}$) & 78.58($\Delta \textcolor{red}{2.42}$)  \\
Dermoscopy             & 58.01($\Delta \textcolor{red}{10.01}$) & 55.28($\Delta \textcolor{red}{11.38}$) & 83.86($\Delta \textcolor{red}{6.82}$)  & 87.69($\Delta \textcolor{red}{4.34}$)  \\
Endoscopy              & 39.94($\Delta \textcolor{red}{13.84}$) & 56.92($\Delta \textcolor{red}{10.64}$) & 57.17($\Delta \textcolor{red}{20.56}$) & 60.34($\Delta \textcolor{red}{12.13}$) \\
Fundus                 & 33.67($\Delta \textcolor{red}{28.50}$)  & 22.99($\Delta \textcolor{red}{27.93}$) & 62.57($\Delta \textcolor{red}{14.62}$) & 76.86($\Delta \textcolor{red}{6.51}$)  \\
Histopathology         & 36.55($\Delta \textcolor{red}{31.92}$) & 79.96($\downarrow \textcolor{green}{0.20}$)  & 79.70($\Delta \textcolor{red}{7.99}$)  & 76.89($\Delta \textcolor{red}{4.31}$)  \\
Microscopy             & 44.92($\Delta \textcolor{red}{15.05}$)  & 78.98($\downarrow \textcolor{green}{0.55}$)  & 70.27($\Delta \textcolor{red}{13.63}$) & 60.83($\Delta \textcolor{red}{13.50}$) \\
Ultrasound             & 15.89($\Delta \textcolor{red}{14.51}$)  & 15.81($\Delta \textcolor{red}{19.85}$) & 55.05($\Delta \textcolor{red}{23.46}$) & 74.81($\Delta \textcolor{red}{10.33}$) \\
X-ray                  & 23.40($\Delta \textcolor{red}{11.04}$)  & 23.12($\Delta \textcolor{red}{16.56}$) & 44.06($\Delta \textcolor{red}{25.84}$) & 64.30($\Delta \textcolor{red}{12.13}$) \\
\bottomrule
\end{tabular}
\end{table}

\textbf{C. Performance Evaluation on Different Modalities.} Figure~\ref{fig3} (b) summarizes the performance of the four methods under the Bbox prompt mode across different modality data. All four methods achieved Dice scores exceeding 70\% in dermoscopy, endoscopy, fundus photography, histopathology, and microscopy. When deploying predicted images at a resolution of $1024\times1024$, SAM outperformed other methods in endoscopy, histopathology, and microscopy modes. We attribute this results to the following factors: 1) These three modalities are derived from 2D datasets and consist of RGB images, which share similarities with natural images. 2) The limited amount of available data for fine-tuning constrained the performance of the fine-tuning method (as indicated by the red lines in the figure, representing the logarithmically transformed mask counts). 3) Larger image resolutions provided more details and resulted in higher prediction performance. In the same resolution setting, our SAM-Med2D significantly outperforms SAM and can effectively handle data from all imaging modalities. Comparing the performance across different modalities directly may be unfair due to the inclusion of different types of objects and variations in data scale. 

Table~\ref{tab3} presents the performance under the point prompt mode. To maintain fairness, we used the same initial point for predictions with SAM (256), FT-SAM, and SAM-Med2D. As observed, the segmentation performance significantly improved with increasing iterations, and SAM achieved improvements of over 10\% across all modalities. What is even more impressive is that our SAM-Med2D, with just a single-point interaction, outperformed the performance of other methods with five-point interactions. This highlights that point prompt, facilitated by large-scale pre-training, can be effectively applied in the field of medical imaging, enabling more efficient interactions compared to Bbox prompt.

\begin{table}[ht]
\renewcommand{\arraystretch}{1.1}
\centering
\caption{Generalization validation on 9 MICCAI2023 datasets, where "*" denotes SAM-Med2D without adapter layer parameters.}\label{tab4}
\small
\setlength{\tabcolsep}{3.5pt}
\begin{tabular}{l|ccc|ccc}
\toprule
\multirow{2}{*}{\textbf{Datasets}} & \multicolumn{3}{c|}{\textbf{Bbox prompt (\%)}} & \multicolumn{3}{c}{\textbf{1 point prompt (\%)}} \\ \cline{2-7} 
                          & \textit{SAM} \cite{08}    & \textit{SAM-Med2D}    & \textit{SAM-Med2D*}    & \textit{SAM} \cite{08}    & \textit{SAM-Med2D}     & \textit{SAM-Med2D*}    \\ \hline
CrossMoDA23\cite{33}               & 78.98      & 70.51      & 84.62       & 18.49       & 46.08       & 73.98       \\
KiTS23 \cite{34}                 & 84.80      & 76.32      & 87.93       & 38.93       & 48.81       & 79.87       \\
FLARE23 \cite{35}                   & 86.11      & 83.51      & 90.95       & 51.05       & 62.86       & 85.10       \\
ATLAS2023 \cite{36}                & 82.98      & 73.70      & 86.56       & 46.89       & 34.72       & 70.42       \\
SEG \cite{37}                       & 75.98      & 68.02      & 84.31       & 11.75       & 48.05       & 69.85       \\
LNQ2023 \cite{38}                   & 72.31      & 63.84      & 81.33       & 3.81        & 44.81       & 59.84       \\
CAS2023 \cite{39}                & 52.34      & 46.11      & 60.38       & 0.45        & 28.79       & 15.19       \\
TDSC-ABUS2023 \cite{40}            & 71.66      & 64.65      & 76.65       & 12.11       & 35.99       & 61.84       \\
ToothFairy2023 \cite{41}            & 65.86      & 57.45      & 75.29       & 1.01        & 32.12       & 47.32       \\ \hline
Weighted average          & 85.35      & 81.93      & 90.12       & 48.08       & 60.31       & 83.41            \\
\bottomrule
\end{tabular}
\end{table}

Furthermore, we noticed that for the histopathology and microscopy modalities, the multi-point interactions of SAM at $1024\times1024$ resolution actually performed worse than the single-point interaction. This may be because the model has already made optimal decisions based on the single-point prompt, and further point prompts aimed at correcting segmentation results may interfere with the model's judgment. This observation also indicates that fine-tuning can enhance the upper limit of SAM's performance.

\textbf{D. Performance Evaluation on Major Organs.} Figure~\ref{fig3} (c) presents the differences in Dice scores between FT-SAM and SAM-Med2D across more than 30 organs. The bar graph represents the results obtained using 5 pts prompt, while the pentagrams indicate the corresponding results with Bbox prompts. SAM-Med2D achieved higher results in 24 organs, with the maximum difference reaching 6.95\%. Furthermore, by observing the experimental results of the same organ under different prompting methods, we can see that the performance gap between 5 points prompt and Bbox prompt is relatively small. This finding suggests that when applying SAM to the medical field, bounding box interaction may not be the most effective approach (as the number of point interactions increases, the absolute advantage of Bbox prompts gradually diminishes). For certain skeletal regions such as ribs, scapulae, and clavicles, the point prompting strategy may be more effective. Overall, these results indicate that models fine-tuned using the Adapter method can achieve exceptional segmentation performance, and point interaction holds significant potential in organ segmentation.
\begin{figure}[ht]
  \centering
  \includegraphics[width=\textwidth]{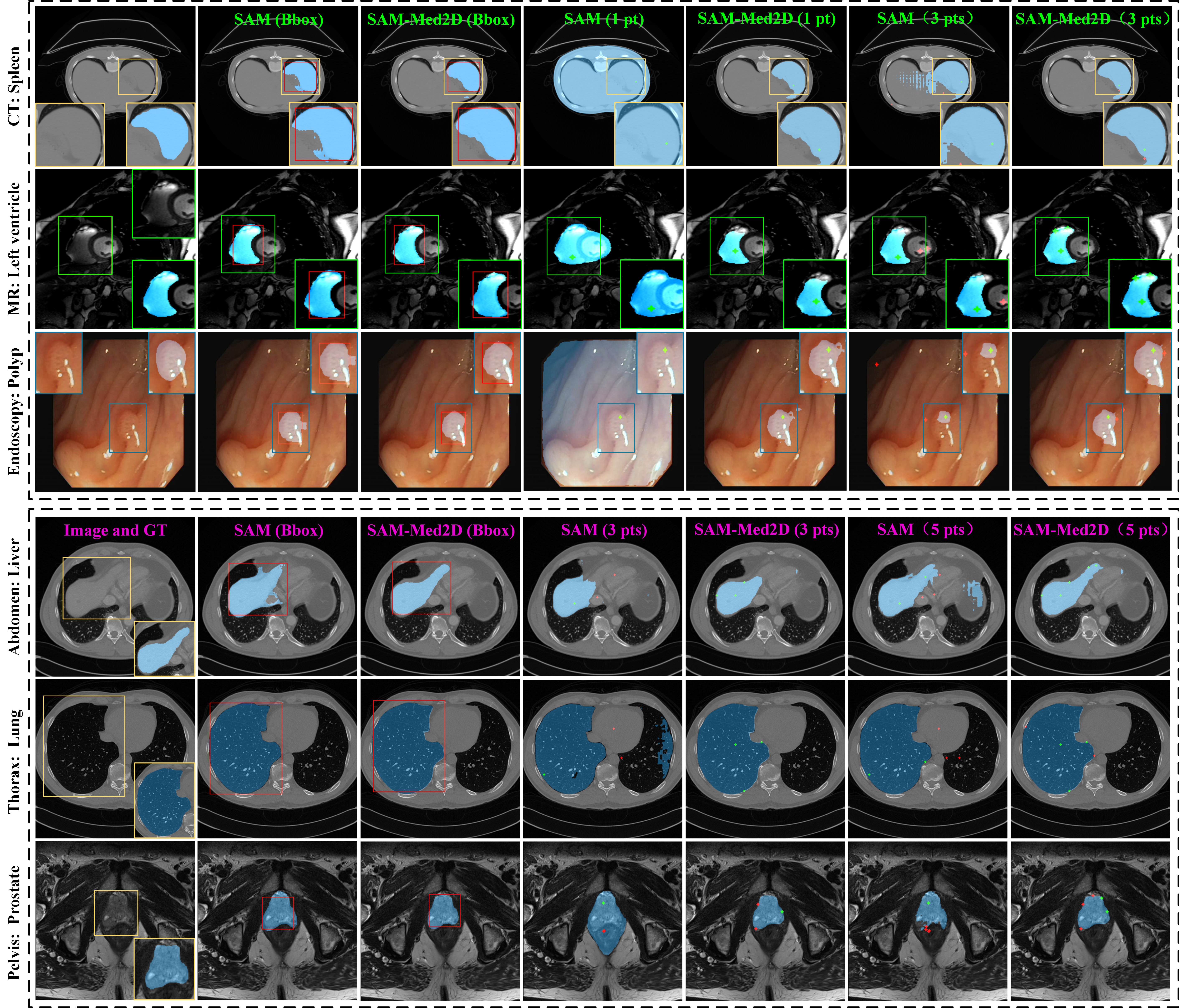}
  \caption{Qualitative comparisons were made between the segmentation results of SAM-Med2D and SAM. The first three rows depict the segmentation results of different modalities, while the last three rows illustrate the segmentation results of different anatomical structures.}\label{fig4}
\end{figure}

\begin{figure}[ht]
  \centering
  \includegraphics[width=\textwidth]{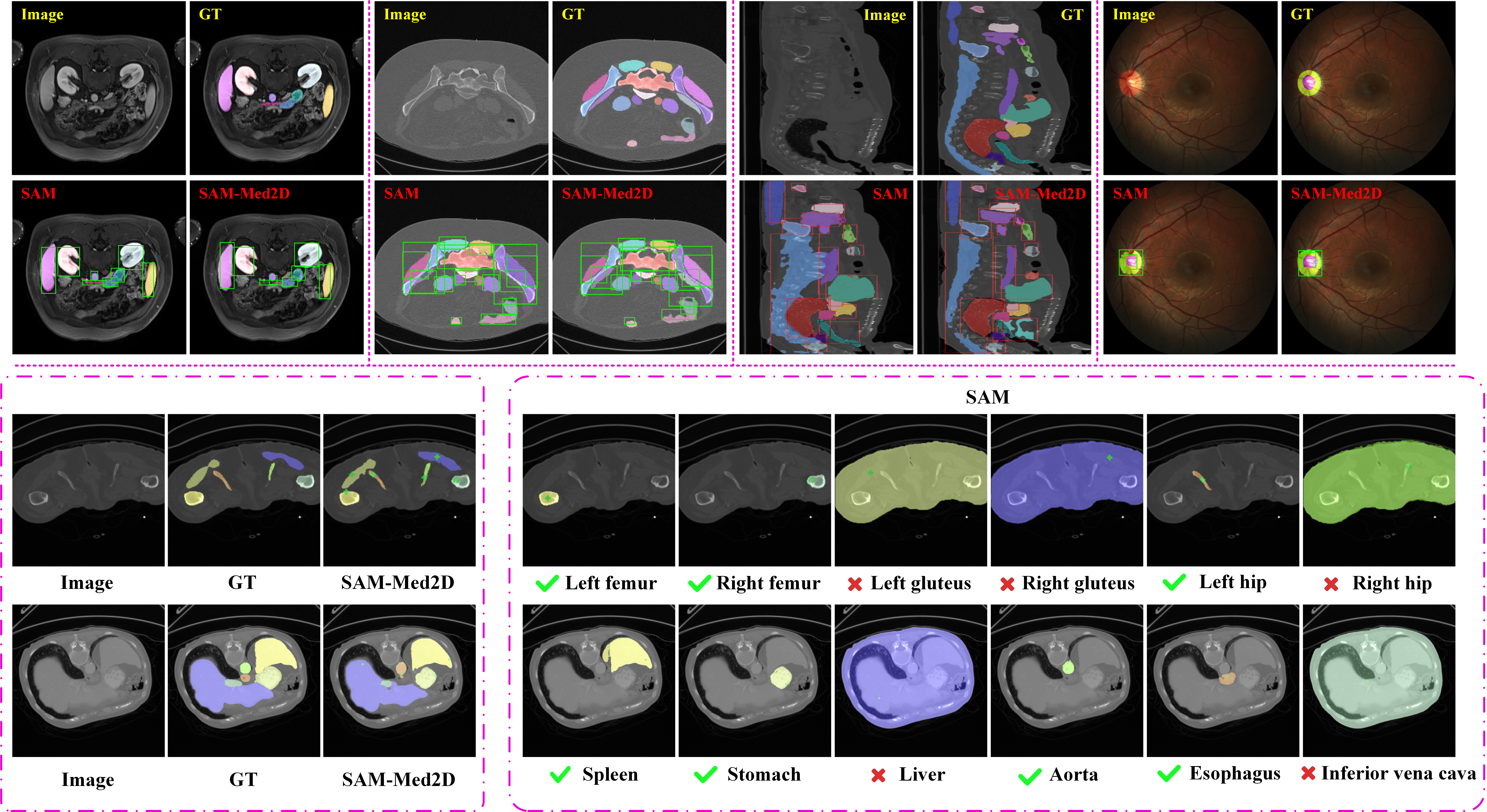}
  \caption{The fusion of segmentation results for multiple target regions within a single image. For clarity of presentation, we visualized only the results of Bbox prompt and 1 point prompt.}\label{fig5}
\end{figure}

\textbf{E. Generalization Evaluation.} To assess the generalization ability of SAM-Med2D, we conducted tests on 9 publicly available datasets, most of which were derived from the MICCAI2023 competition \cite{33,34,35,36,37,38,39,40,41}. To ensure a fair comparison, we uniformly used 256×256 resolution images for testing. Table~\ref{tab4} shows that SAM demonstrated excellent performance under bounding box prompts, with a weighted average Dice of 85.35\%. However, its performance under single-point prompts was not satisfactory (48.08\%). Since the adapter layer parameters are plug-and-play, we tested two scenarios: retaining and removing the adapter layer parameters. When the adapter layer parameters were retained, our SAM-Med2D achieved a Dice score of 81.93\% under Bbox prompts, and the performance improved by 8.19\% when the adapter layer parameters were removed. Additionally, we observed that SAM failed to adapt effectively to point prompt, with the best segmentation performance across the 9 datasets reaching only 51.05\%. In contrast, our SAM-Med2D obtained reasonable segmentation results under point prompt. It is worth noting that when we removed the adapter layer parameters during inference, the performance of SAM-Med2D with 1 pt prompt was very close to that of SAM under Bbox prompt (83.41\% vs. 85.35\%), which saves a significant amount of time and cost for data annotation and analysis. In summary, SAM showed good generalization performance only under bounding box prompts, while our SAM-Med2D achieved better generalization under both prompting modes.

\subsection{Qualitative Comparison}

We qualitatively compared the segmentation masks of SAM-Med2D and SAM. The visual results of SAM are derived from the better of the two resolutions, namely 256×256 or 1024×1024. The first three rows of Figure~\ref{fig4} illustrate the segmentation performance of both methods on three modals. In most cases, the segmentation results indicated by Bbox can locate the target region, but the boundaries in our SAM-Med2D visual results are clearer and closer to ground truth. In the case of 1 pt prompt, SAM struggles to locate the target region, leading to significant discrepancies between the segmentation results and the expected outcome.

The last three rows depict the segmentation results of both models on liver, lung, and prostate organs. For Bbox prompt mode, both methods can generate masks of similar quality. By observing the results of 3 pts and 5 pts prompt, we can see that more point prompt result in better segmentation outcomes. Among models with the same number of point prompt, SAM-Med2D can describe the target region better than SAM, implying that SAM-Med2D requires fewer interactive operations and less time to achieve the desired results. This is advantageous for data annotation or pseudolabel generation. We attribute this phenomenon to SAM-Med2D acquiring domain-specific knowledge related to the medical imaging field through learning from large-scale datasets. This aligns with the motivation of this paper, which is to establish the foundation for SAM to move towards robust and reliable medical image segmentation.

Figure~\ref{fig5} shows the results of merging multiple target regions within the same image. When the target boundaries are clear, there are subtle visual differences between SAM and our SAM-Med2D. In other cases, SAM-Med2D can achieve precise segmentation of parts that are difficult for the human eye to identify. On the other hand, in the scenario of 1 pt prompt, SAM often fails on many organs and struggles to locate the target region. This once again demonstrates that fine-tuning SAM on large-scale data can lead to better domain transferability.

\section{Discussion and Conclusion}

In this study, we obtain SAM-Med2D by fine-tuning a SAM on a large-scale medical image dataset, which is able to significantly improve various medical image segmentation tasks. We employed two explicit prompts strategies to generate masks for quantitative and qualitative comparisons. At an equal resolution, only the fine-tuned mask decoder (FT-SAM) achieved an improvement of 11.93\% in the Bbox prompt mode, while the fully fine-tuned SAM-Med2D achieved a 17.67\% improvement. Surprisingly, our approach demonstrated overwhelming superiority in the 1 pt prompt (18.94\% vs. 70.01\%). Furthermore, SAM-Med2D exhibited excellent generalization capabilities in both prompt modes, indicating its practical value in the medical field.

We conduct a comprehensive evaluation of the model from different dimensions of the data. From an anatomical perspective, at a resolution of 1024×1024, SAM had advantages over FT-SAM in the chest, abdomen, and other regions, SAM-Med2D outperformed all other methods in overall segmentation performance. Regarding different modalities, SAM demonstrated good generalization when the target modality data resembled natural image attributes. We compared the two fine-tuning methods on more than 30 major organs, and our SAM-Med2D achieved better results on 24 organs, with a maximum improvement of 6.95\% compared to FT-SAM. Additionally, our generalization experiments on 9 publicly available datasets demonstrated strong domain transferability of models pretrained on large-scale datasets. While the Bbox prompt always outperformed the 1 pt prompt, adding more points significantly improved the segmentation results, surpassing even the Bbox mode. When using the point prompt mode, SAM-Med2D was able to generate the desired masks more quickly, even outperforming other methods using the Bbox prompt mode.

However, further optimization is required for more robust medical image segmentation in the future. From the qualitative segmentation results, for complex shapes/boundaries, small size or low contrast objects, there is still room for improvement in the segmentation results produced by different prompt modes. Establishing relevant optimization strategies in the future may enhance the segmentation results, such as setting window width for different organs and designing boundary loss for interactive segmentation. In addition to the prompt strategies used in this paper, natural language can serve as another common form of user interaction in medical image segmentation, but currently, there is a lack of relevant datasets. This is a direction we actively pursue, aiming to equip SAM-Med2D with natural language understanding capabilities in the medical domain to meet diverse user needs. While we have trained SAM-Med2D on over 19.7M masks, there is still a gap compared to the training data and resolution of SAM. This results in SAM-Med2D being effective in handling common organs or lesions but lacking the "everything" capability in the medical domain. We plan to generate a larger quantity and wider variety of high-quality masks through data engines, enabling SAM-Med2D to truly segment all types of medical images.

In summary, this work fine-tuned the SAM on a large-scale medical image dataset to adapt it to the medical image domain. SAM-Med2D achieved satisfactory performance improvements and generalization capabilities. Our code and pre-trained models will be made available for researchers, and we hope this work benefits researchers in the field of medical computer vision, providing insights and opportunities for future research and improvements.


%


\end{document}